# An affective and adaptive educational robot

CRISTINA GENA, CS Dept., University of Turin, Italy

ALBERTO LILLO, CS Dept., University of Turin, Italy

CLAUDIO MATTUTINO, CS Dept., University of Turin, Italy

ENRICO MOSCA, CS Dept., University of Turin, Italy

In this paper we present an educational robot called Wolly, designed to engage children in an affective and social interaction. Indeed, we are now focusing on its role as an educational and affective robot capable of being controlled by coding instructions and at the same time interacting verbally and affectively with children by recognizing their emotions and remembering their interests, and adapting its behavior accordingly.

## 1 INTRODUCTION

Educational robotics [1] is a simple and practical approach, used in schools, to make STEM subjects learn in a fun and interesting way through robots from the earliest years of school . The educational robotics approach not only facilitates socialization since leverages the workgroup, but also helps to stimulate the curiosity and logic actuating the so-called computational thinking [9].

Educational robotics seems to be a very useful approach also with regard to inclusion, since it can serve as a" *tool for knowledge construction and as an assistive tool for students who have problems in specific fields, or ER may be used to change students' attitudes to learning—class culture—allowing everyone to be accepted and involved*" [14]. For example, children with cognitive and relational disabilities, as for instance autistic children, may acquire a sense of control that they normally do not have [13], [6]. Some robots can communicate through different channels, perhaps showing images while talking or gesticulating, making the interaction suitable and inclusive even for different types of disabilities. In this paper we present an educational robot, designed both to engage children in an affective and social interaction, and to be programmable by the children also in its social and affective behavior. The paper is organized as follows: Section 2 presents the robot and its main features, Section 3 describes the robot's hardware and software architecture, Section 4 presents current and future developments, and Section 5 concludes the paper.

## 2 THE WOLLY ROBOT

At the end of 2017, in our HCI lab we carried out a co-design activity with children aimed at devising an educational robot called Wolly [3], [8]. The main goal of the robot is behaving as an affective peer for children: it has to be able not only to execute a standard set of commands, compatible with those used in coding and educational robotics, but also to interact both verbally and affectively with students. Currently, Wolly can be controlled by means of a standard visual block environment, Blockly[1], see Fig. 1, which is well known to many children with some experience in coding. However, we also have a simpler set of instructions, see Fig. 1, we tested with children (as described in [2]), so that younger children can use basic commands to control its behavior.

---

[1] https://developers.google.com/blockly/

We are now focusing on its role as an educational and affective robot capable of being controlled by coding instructions and at the same time interacting verbally and affectively with children by recognizing their emotions and remembering their interests, and adapting its behavior accordingly.

## 3 HARDWARE AND SOFTWARE ARCHITECTURE

### 3.1 Hardware architecture

Currently we have developed a second release of the robot (as far as the first release is concerned see [3]). We had the requirement of transforming Wolly into a low cost robot that everyone could build regardless of its electronic and technical skills. The idea is to convert the old Arduino-based robot architecture (see details in [8]) in a all-in-one device managed by a Raspberry Pi, with a sort of WollyOS (Wolly Operating System) easy to download and update, also for nontechnical users, as teachers. In more details, the list of the main hardware components we are now using is the following: Raspberry Pi model 3 or higher; Screen 5 inch HDMI LCD - for Raspberry Pi 3B + / 4B; Waveshare Motor Driver Hat for Raspberry Pi - Two DC Motors I2C Interface 5V 2 gear motors 5V; Tracks Baoblaze in technopolymer/Tamiya 70100; Bluetooth Speaker JBL Go2; Raspberry Pi Camera 5MP 1080P; Power bank 2000mAh or higher.

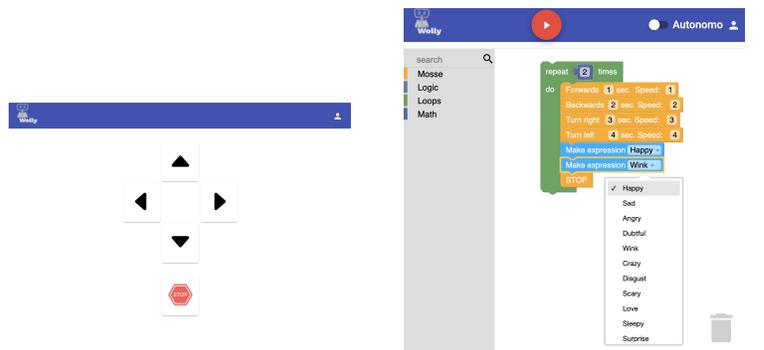

(a) The robot can be moved with the 4 arrows in the page, and there is s also a stop button to stop it.

(b) The Blockly Page.

Fig. 1. Main pages of Wolly Site

### 3.2 Software architecture

As far as the software components of the second release are concerned, Wolly is now controlled both from a desktop web-based interface and from a smartphone/tablet application. The responsive web-based application is developed in Angular [2] and the commands for the robot are sent and read in Python. In the future there will be a mobile application developed in Flutter[3] presenting a webview showing the web pages through which the user can take control of Wolly both on iOS and Android.

---

[2] https://angular.io/docs
[3] https://flutter.dev/

More in detail, we are using the following software components (see also Figure 2).

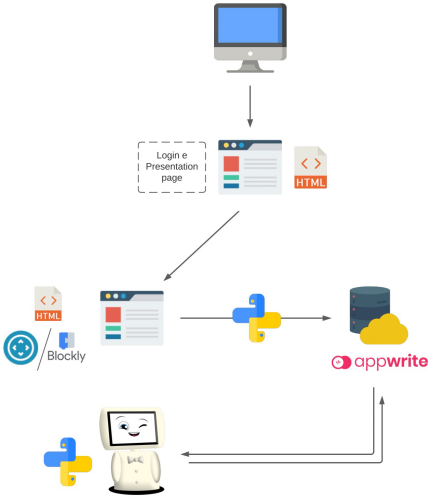

Fig. 2. Software architecture

Appwrite[4]. Appwrite is an end-to-end back-end server that is aimed to abstract the complexity of common, complex, and repetitive tasks required for building a modern app. Appwrite provides the developer with a set of tools to build apps a lot faster and in a much more secure way. Appwrite is used to build different services such as user authentication and account management and real-time database. Appwrite can run on any operating system and coding language. The communication between Wolly and the other components takes place through the Appwrite database. Every action is written on the database and the robot is always listening for changes (instructions). Whenever they occur Wolly will execute them, and when the set of commands will be finished, the database will be reset and ready for receiving a new set of commands [5].

Blockly. This library adds an editor representing the coding concepts as interlocking graphical blocks to represent code concepts like variables, logical expressions, loops and, thanks to the customized blocks, to move the robot. As outputs, it returns syntactically correct Python code that is eventually sent to the database. Blockly is an open-source library, so it's very easy to create and customize your own block. Indeed, via the *Blockly Developer Tools* [5] it is possible to design a special block from scratch by choosing the colors, the shape, how it will be translated and in which language, in our case we have chosen Python. In our application, we designed six customized blocks: Move Forward; Move Right; Move Left; Move Backward; Stop Make an Expression.

The "Make an Expression" block has a drop-down menu for choosing one of the eleven different available expressions the robot is able to express, see Fig. 1.

---

[4] https://appwrite.io
[5] https://blockly-demo.appspot.com/static/demos/blockfactory/index.html

## 4 CURRENT AND FUTURE WORKS

### 4.1 Emotion Recognition

We are currently working to to give Wolly some sort of social behaviour, thus we decided to start adding it emotion recognition functionalities. For this purpose a we created a Neural Network, in particular we used a Convolutional Neural Network (CNN). The neural network not only recognize emotions, but it is able to give a prediction based on the user's body and context. Since the overall expression of emotion may vary across contextual situations, we decide to use the Emotic Dataset[6] [11, 12], which uses also contextual information in recognizing emotions.

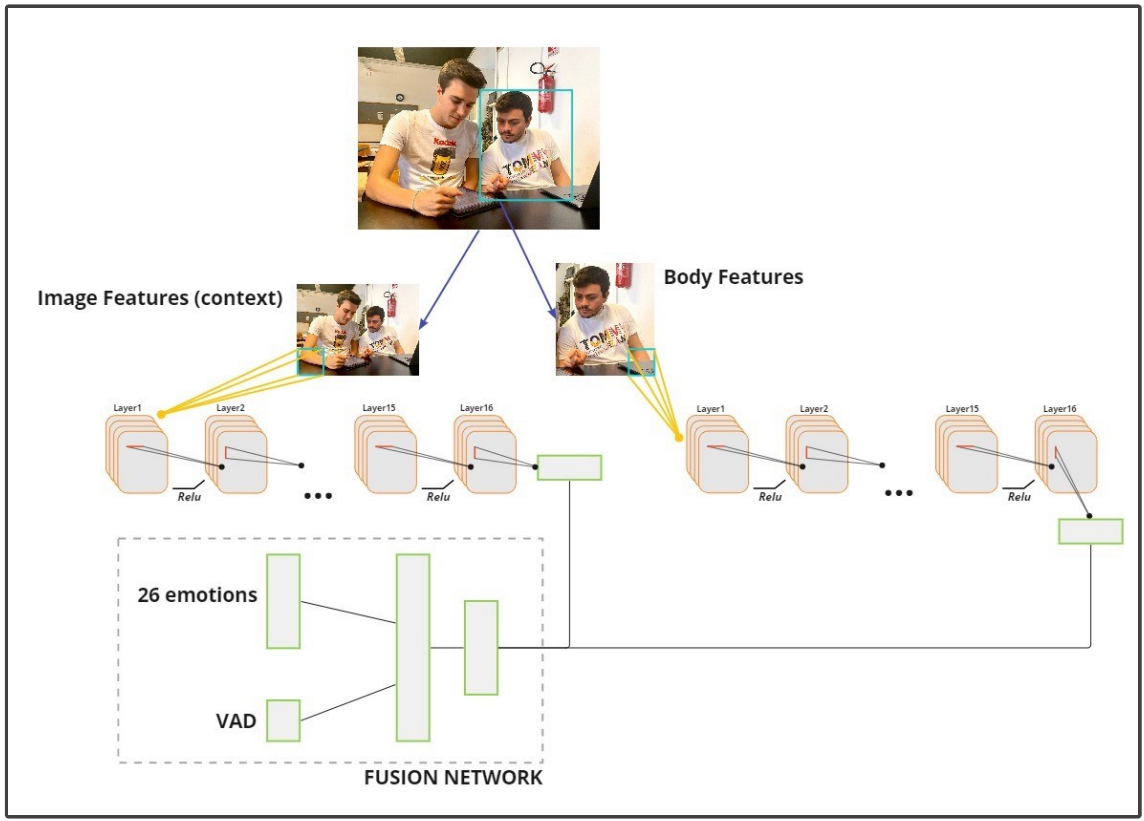

Fig. 3. The network model consists of two modules that extract the features and a fusion network for estimating emotions and VAD

*4.1.1  The CNN network.* The CNN uses two feature extraction modules to extract features over an image, these are then used by a third module (Fusion Network) to predict the dimensions (valence, arousal and dominance) and the emotion categories. Each extraction modules network consists of 16 convolutional layers with 1-dimensional kernels alternating between horizontal and vertical orientations.

---

[6] http://sunai.uoc.edu/emotic/

- Body feature extraction: analyze the visible part of the body and extract body-related features (i.e. face and head aspects, pose or body appearance). This module is pre-trained with ImageNet [7].
- Image feature extraction: analyze the whole image and extract scene-context features (i.e. objects present in the scene, dynamics between people in the scene, etc..). This module is pre-trained with Places dataset [8].
- Fusion Network: combines the features extracted by the other two modules and predict the emotion categories and VAD.

*4.1.2    Training and results.* We trained the network with both the training set and validation set for 15 epoch (Fig.4a). The network has been trained end-to-end, learning the parameters jointly using *stochastic gradient descent with momentum*. The loss function is a weighted combination of two separate losses, one of the emotions categories and the other one of the VAD values. A prediction $\hat{y}$ is composed by the prediction of each of the 26 emotion categories (*Cat*) and the 3 VAD values (*Cont*).

$$\hat{y} = (\hat{y}_{cat}, \hat{y}_{cont})$$

In particular, Fig.4a shows the loss, after each one of the 15 epoch, for the training set (*loss*) and validation set (*validation loss*) obtained combining:

- *Cat Loss (Category loss)*
- *Cont Loss (Continuous loss)*

In Fig.4b are shown the precision values for each category (emotions) after the testing.

Furthermore we obtained the following precision values for:

- *Valence*: 0.70991
- *Arousal*: 0.87199
- *Dominance*: 0.90254

After the testing we obtained a Mean Average precision for the predictions of the Emotions Category and the Mean Error of the VAD values, respectively of *0.27* and *0.83*.

---

[7] https://www.image-net.org/
[8] http://places2.csail.mit.edu/

```
train  context  (23266, 224, 224, 3) body (23266, 128, 128, 3) cat  (23266, 26) cont (23266, 3)
val  context   (3315, 224, 224, 3) body (3315, 128, 128, 3) cat  (3315, 26) cont (3315, 3)
train loader  448 val loader  64
completed preparing context model
completed preparing body model
starting training
epoch = 0 loss = 58975.2319 cat loss = 28637.7351 cont_loss = 89312.7287
epoch = 0 validation loss = 5406.7201 cat loss = 5456.3504 cont loss = 5357.0897
epoch = 1 loss = 43887.2341 cat loss = 17921.4714 cont_loss = 69852.9966
epoch = 1 validation loss = 5296.8695 cat loss = 5442.3076 cont loss = 5151.4314
epoch = 2 loss = 43530.0574 cat loss = 17860.4012 cont_loss = 69199.7135
epoch = 2 validation loss = 5080.4812 cat loss = 5392.0961 cont loss = 4768.8663
epoch = 3 loss = 43386.1295 cat loss = 17887.2132 cont_loss = 68885.0457
epoch = 3 validation loss = 5033.9900 cat loss = 5420.8421 cont loss = 4647.1378
epoch = 4 loss = 43305.4891 cat loss = 17833.3195 cont_loss = 68777.6587
epoch = 4 validation loss = 5039.9719 cat loss = 5305.5835 cont loss = 4774.3604
epoch = 5 loss = 43225.7772 cat loss = 17841.9623 cont_loss = 68609.5920
epoch = 5 validation loss = 5604.0765 cat loss = 5567.3226 cont loss = 5640.8304
epoch = 6 loss = 43252.1920 cat loss = 17823.6520 cont_loss = 68680.7319
epoch = 6 validation loss = 5270.9732 cat loss = 5364.9560 cont loss = 5176.9906
epoch = 7 loss = 42823.2062 cat loss = 17753.2706 cont_loss = 67893.1416
epoch = 7 validation loss = 5163.3261 cat loss = 5428.4211 cont loss = 4898.2310
epoch = 8 loss = 42764.6032 cat loss = 17767.2745 cont_loss = 67761.9318
epoch = 8 validation loss = 5126.3762 cat loss = 5408.2318 cont loss = 4844.5207
epoch = 9 loss = 42688.1876 cat loss = 17753.2455 cont_loss = 67623.1296
epoch = 9 validation loss = 5094.4931 cat loss = 5396.0250 cont loss = 4792.9613
epoch = 10 loss = 42668.1838 cat loss = 17760.8847 cont_loss = 67575.4828
epoch = 10 validation loss = 5152.6034 cat loss = 5415.0499 cont loss = 4890.1569
epoch = 11 loss = 42589.8462 cat loss = 17726.4400 cont_loss = 67453.2524
epoch = 11 validation loss = 5061.3120 cat loss = 5385.6304 cont loss = 4736.9936
epoch = 12 loss = 42453.2270 cat loss = 17704.9001 cont_loss = 67201.5538
epoch = 12 validation loss = 5171.4250 cat loss = 5418.8587 cont loss = 4923.9912
epoch = 13 loss = 42313.1187 cat loss = 17672.9335 cont_loss = 66953.3038
epoch = 13 validation loss = 5104.1341 cat loss = 5387.2423 cont loss = 4821.0261
epoch = 14 loss = 42186.1946 cat loss = 17693.1449 cont_loss = 66679.2442
epoch = 14 validation loss = 5075.2083 cat loss = 5384.3507 cont loss = 4766.0657
completed training
```

```
starting testing
completed testing
saved mat files
Category          Affection 0.26058
Category              Anger 0.06201
Category          Annoyance 0.10618
Category       Anticipation 0.92964
Category           Aversion 0.09045
Category         Confidence 0.73932
Category        Disapproval 0.10081
Category      Disconnection 0.28152
Category       Disquietment 0.15382
Category      Doubt/Confusion 0.14854
Category      Embarrassment 0.06390
Category         Engagement 0.96377
Category             Esteem 0.20318
Category         Excitement 0.71706
Category            Fatigue 0.07537
Category               Fear 0.06656
Category          Happiness 0.69400
Category               Pain 0.05700
Category              Peace 0.21237
Category           Pleasure 0.38222
Category            Sadness 0.07102
Category        Sensitivity 0.06361
Category          Suffering 0.06635
Category           Surprise 0.11121
Category           Sympathy 0.25966
Category           Yearning 0.10405
Mean AP 0.26862
Continuous    Valence 0.70991
Continuous    Arousal 0.87199
Continuous  Dominance 0.90254
Mean VAD Error 0.82815
saved thresholds
```

(a) Loss values during the training.

(b) Precision of each category and VAD during the testing.

Fig. 4. Training and test values

From the trained network, we retrieve a list of the predicted categories. For example in the picture 5 the predicted emotions are: engagement, excitement, happiness and pleasure

Every prediction also gives 3 Continuous Emotion Dimensions (range [0-1]):

- *Valence (V)*: measures how positive or pleasant an emotion is. For example if the valence is high, tending to 1, the feeling is pleasant, so the person is happy;
- *Arousal (A)*: measures the agitation level of the person, ranging from non-active/calm to agitated/ready to act;
- *Dominance (D)*: measures the control level a person feels about the current situation.

We will use them combined with the predicted emotions to infer the overall feeling of a children talking to Wolly.

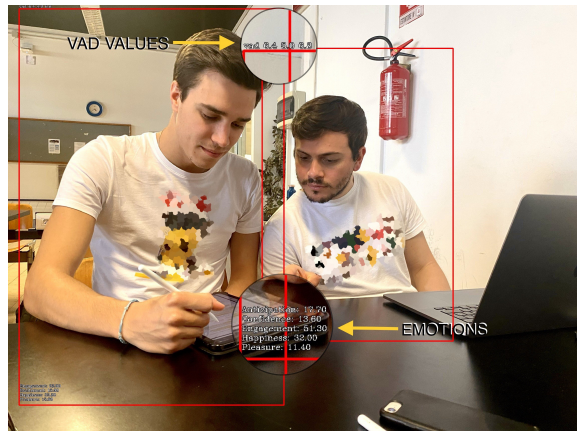

Fig. 5. An example of how the image is analyzed by the network

*4.1.3* Wolly API. We are now working for having the Emotion Recognition module be available via Rest APIs. We created a server that run the Pyhton programs and make predictions over the images that it receives and and returns the them in response. Thus the robot will work as a client that capture the images of the user on which an emotion is to be identified, send the frames to the server and waits for its response about the recognized emotions.

The data of the analyzed image are returned as a JSON containing the details (emotions and VAD values) for each person in the scene as shown in listing ??. This results refers to Fig. 5

```
{
    "data" : {
        "0" : {
            "emotions" : {
                "Engagement" : "53.10",
                "Excitement" : "15.40",
                "Happiness" : "35.10",
                "Pleasure" : "15.00"
            },
            "vad" : [
                6.384382724761963,
                4.854542255401611,
                6.836206436157227
            ]
        },
        "1" : {
```

```json
        "emotions": {
            "Anticipation":"17.70",
            "Confidence":"13.60",
            "Engagement":"51.30"
            "Happiness":"32.00",
            "Pleasure":"11.40"
        },
        "vad": [
            6.392988204956055,
            4.880496501922607,
            6.800170421600342
        ]
      }
   }
}
```

## 4.2 Face Recognition

For biometric recognition we used the FaceRecognition library [9], which is one of the most used library for recognizing and manipulating faces. Thanks to this Python library the robot can recognize people, so that it can recognize who already knows and react accordingly. If the person in front of the robot is unknown, Wolly will interact with her/him asking questions about the name, age and, if the user agrees, it will take and save a picture of her/him. This information will be then used to recognize and greet the user, and to produce more accurate and personal dialogues between the human and the robot.

## 4.3 Ontology

We decided to enrich Wolly with a knowledge base focused on movie's and cartoon's domain, and to achieve this goal we created an OWL-based ontology. We will use this knowledge base in order to extend the robot's capability of autonomously talking and interacting via natural dialogues with the children, by answering questions about movies and cartoons.

---

[9] https://github.com/ageitgey/face_recognition

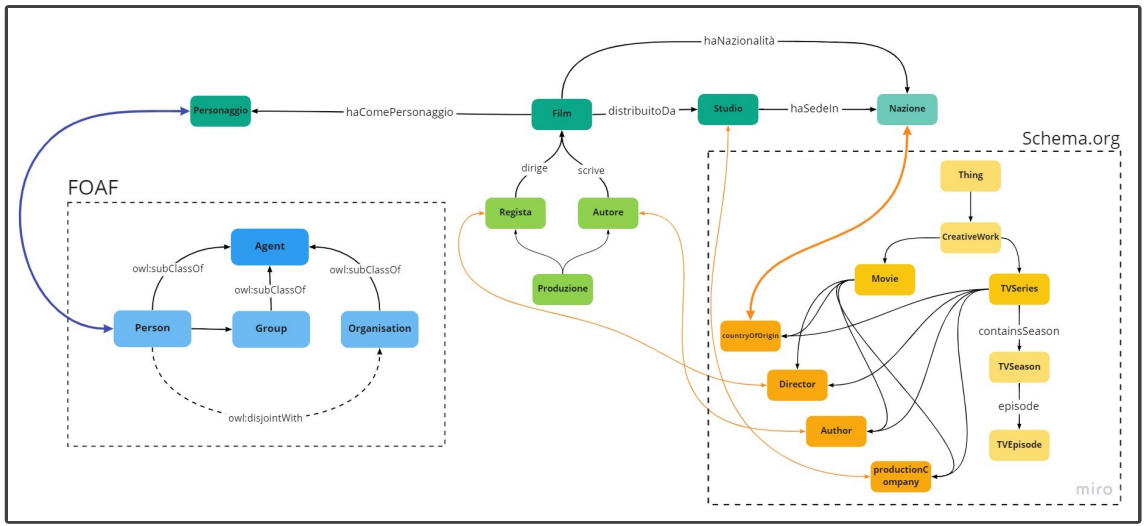

Fig. 6. The ontology

The ontology has been aligned with external and shared ontologies, in particular we used FOAF and Schema.org.

*4.3.1* FOAF. [10] is used for the representation of characters, so an important alignment adopted is the FOAF (Friend of a Friend) resource. FOAF is an ontology designed to describe people (real or imaginary), their activities and relationships with other people and objects.

FOAF can be seen as a descriptive vocabulary expressed in the Resource Description Framework (*RDF*) and defined using Web Ontology Language (*OWL*). Wolly can therefore use FOAF, for example, to search for all the characters who have starred in a movie, or all the characters who act with another character, because FOAF is also designed to define relationships between people.

*4.3.2* Schema.org. [11] is based on RDF schema, and is currently used to describe the movie classes, sub-classes and properties. Schema.org offers vocabularies covering entities, relationships between entities and actions, and can easily be extended through a well-documented extension model.

### 4.4 Natural language interaction

We are now also working on speech recognition and generation thanks to the integration of Google Cloud Text-toSpeech [12] and Google Speech to Text API[13]. We would also integrate in the near future the dialogue module described in Gena et al. [7] that uses algorithms for recognizing groups of synonyms, which can be easily customized by the user and also allows easy management of the user / robot dialogue in a way similar to Aldebaran's QiChat[14].

---

[10] http://xmlns.com/foaf/spec/
[11] https://schema.org/
[12] https://cloud.google.com/text-to-speech
[13] https://cloud.google.com/speech-to-text

As future work, we will integrate the robot's dialogue with this knowledge base described above, in order to have the robot able to move and reason on the ontology, and thus enriching its dialogue's strategies. For instance, the taxonomic structure of the ontology could allow Wolly to drive the focus of the conversation to related topics or to more general or specific topics, and, in general, it could improve its capability to manage the conversation and disambiguate the input from the user

In addition to classic programming tasks, the robot may in the future be programmed as a social robot. In addition to movements, the user may make the robot express emotions and make it say things, programming in this way its dialogue and its social and affective behavior.

## 5 CONCLUSION

In the future we want to make the robot be able to carry out a basic dialogue interaction linked to both the context of the coding exercises and also to its general knowledge about the world and about itself, as well as about specific topics, as the ones linked to the ontology above described, i.e. movies and cartoons.

We would also to make the robot able to adapt its behavior according to the user's interests (discovered through conversation) to the user's context and perceived emotions. In fact, as a long-term goal we plan to enrich the robot with a user modeling component, which will keep track of past interactions with the user, will reason about her/him on the basis of her/his features, skills and emotions recorded during past interactions, and will be able to better assist the user in her/his choices [10]. Finally we will carry out an evaluation with real users, according to the methodologies we experimented in the past [4].

---

[14] http://doc.aldebaran.com/2-5/naoqi/interaction/dialog/aldialog_syntax_overview.html